\documentclass[runningheads]{llncs}
\usepackage[T1]{fontenc}
\usepackage{graphicx}
\usepackage{amsfonts, amsmath, bm}  
\usepackage{algorithmic, algorithm} 

\begin{document}
\title{Environment-Centric Active Inference}
%
%
\author{Kanako Esaki\inst{1} \and
Tadayuki Matsumura\inst{1} \and
Takeshi Kato\inst{2} \and\\
Shunsuke Minusa\inst{1} \and
Yang Shao\inst{1} \and
Hiroyuki Mizuno\inst{1}}
\authorrunning{K. Esaki et al.}
%
\institute{Hitachi, Ltd., Tokyo, Japan \and
Kyoto University, Kyoto, Japan}
\maketitle              
\begin{abstract}
To handle unintended changes in the environment by agents, we propose an environment-centric active inference EC-AIF in which the Markov Blanket of active inference is defined starting from the environment. In normal active inference, the Markov Blanket is defined starting from the agent. That is, first the agent was defined as the entity that performs the ``action'' such as a robot or a person, then the environment was defined as other people or objects that are directly affected by the agent's ``action,'' and the boundary between the agent and the environment was defined as the Markov Blanket. This agent-centric definition does not allow the agent to respond to unintended changes in the environment caused by factors outside of the defined environment. In the proposed EC-AIF, there is no entity corresponding to an agent. The environment includes all observable things, including people and things conventionally considered to be the environment, as well as entities that perform ``actions'' such as robots and people. Accordingly, all states, including robots and people, are included in inference targets, eliminating unintended changes in the environment. The EC-AIF was applied to a robot arm and validated with an object transport task by the robot arm. The results showed that the robot arm successfully transported objects while responding to changes in the target position of the object and to changes in the orientation of another robot arm.

\keywords{robot  \and Markov Blanket \and embodied agent \and object transport.}
\end{abstract}
\section{Introduction}
Active inference, which explains the intelligence of living organisms, has increased the intelligence of various agents. According to the free energy principle, which is the basis of active inference, living organisms minimize their free energy by either changing their actions for sampling the environment or by changing their perceptions for inferring environmental states~\cite{FEP:Friston:2006,FEP:Friston:2010,FEP:McGregor:2015}. The unique feature of this principle is that changing action, i.e., active inference, is also explained in a unified manner~\cite{AIF:Friston:2017,AIF:Parr:2022}. This sensorimotor contingency, in which perception and action are treated in a unified manner, is compatible with agents such as robots because it eliminates actions that are unrelated to perception. Thus, active inference has been implemented in various agents and has contributed to generate intelligent actions~\cite{AIFtoAgent:Catal:2020,AIFtoAgent:Fountas:2020,AIFtoAgent:Friston:2020,AIFtoAgent:Horii:2021,AIFtoAgent:Lanillos:2021,AIFtoAgent:Millidge:2020,AIFtoAgent:Pezzato:2023,AIFtoAgent:Sajid:2021,AIFtoAgent:Ueltzhoffer:2018,AIFtoAgent:Esaki:2021,AIFtoAgent:Esaki:2024}.

Defining the agent and the environment in active inference has been left to researchers~\cite{AIFtoEmbodiment:Buckley:2024,AIFtoEmbodiment:Leo:2016,AIFtoEmbodiment:Oliver:2022,AIFtoEmbodiment:Van:2024,AIFtoEmbodiment:Çatal:2020}. In these studies, especially in systems with an embodiment such as a robot, the implicit guideline has been to define the robot as the agent and the surroundings as the environment. Defining the agent and the environment implies designing Markov Blanket of active inference. As long as the defined environment is maintained, the agent will outperform the human in some cases. However, changes in the environment that cannot be directly changed by the agent or in the presence of other agents can occur. The implicit guideline for designing Markov Blanket would be unable to respond to such changes in the environment that the agent does not intend.

Design strategies for Markov Blanket are required that can accommodate unintended changes in the environment. It is not an implicit guideline, but a specific strategy that is independent of the system configuration, such as the robot and its surroundings. What are agents and environments? The design strategy should answer this essential question. Changes in the environment that are not intended by the agent originate from assuming an entity corresponding to the agent (e.g., a robot) and defining its surroundings as the environment. Instead of assuming the entity corresponding to the agent, considering all of the world as the environment, the changes in the environment must be under the intention of the agent. Therefore, defining a Markov Blanket based on the environment by assuming that everything in the world is the environment would allow the agent to respond to changes in the environment.

\begin{figure}
    \includegraphics[width=\textwidth]{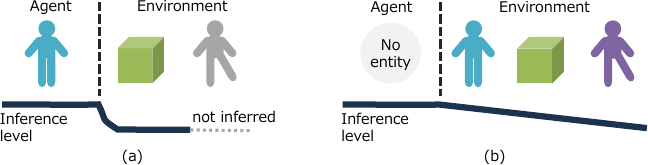}
    \caption{Concept of (a) conventional Markov Blanket (b) Markov Blanket of EC-AIF.} \label{ConceptMB}
\end{figure}

We propose an environment-centric active inference EC-AIF that designs Markov Blanket based on the environment. The concept of Markov Blanket is shown in Fig.~\ref{ConceptMB}. The conventional Markov Blanket has defined the agent as the entity that performs the ``action'', such as a robot or a person, and the environment as the other people or objects, as shown in Fig.~\ref{ConceptMB}(a). The state of the agent is fully known. In addition, the state of the environment within the definition is inferred with high precision, but the state of the environment outside the definition is not inferred at all. The proposed EC-AIF is completely different, as shown in Fig.~\ref{ConceptMB}(b): there is no entity corresponding to an agent. In addition to the people and objects that have been considered the environment, the environment includes the entities that perform ``actions,'' such as robots and people. The state of the environment is inferred with varying accuracy. The EC-AIF was applied to a robot arm and demonstrated in an object transport task. The robot arm adapted to changes in the environment and successfully transported the object.

\section{Method}
\subsection{Free Energy Principle and Active Inference}
Living organisms are said to follow the Free Energy Principle (FEP). Living organisms repeat a process of perception and action. Perception is the process of acquiring an observation $o$ from the environment and inferring a hidden state $s$ of the environment. Action is the process of inferring the appropriate policy $\pi$ and acting on the environment. The hidden state $s$ and policy $\pi$ are inferred so that the free energy becomes smaller, using the generative model $p(o,s,\pi)$ of the environment that living organisms have. One important feature of the FEP is that actions are also regarded as inferences of policy, i.e., active inference. FEP handles perception and action as a unified ``inference.''

Inference assumes that the categories of perception and action are predefined. Perception and action are represented by an observation $o$, a hidden state $s$, and a policy $\pi$, which are treated as probability distributions. Accordingly, defining the probability variables for each of observation $o$, hidden state $s$, and policy $\pi$ is equivalent to defining the categories of perception and action. For example, consider the case where the observation variable is the retinal image, the hidden state variable is the position of the target object, and the policy variable is the direction of eye movement. In this case, perception is to infer the position of the target object based on the retinal image, and action is to infer the appropriate direction to move the eye based on the position of the target object. The set of these probability variables is called the Markov Blanket. Markov Blanket determines the categories of perception and action.

\subsection{Markov Blanket}
Living organisms are considered to adaptively select their Markov Blankets. The Markov Blanket commonly imagined is the boundary between the living organism's body and its surroundings. However, this is not the only Markov Blanket of a living organism. Living organisms have Markov Blankets as hierarchical structures, such as the boundary between organs and their surroundings, and between cells and their surroundings~\cite{MarkovBlankets:Friston:2019,MarkovBlankets:Kirchhoff:2018,MarkovBlankets:Palacios:2020,MarkovBlankets:Pearl:1988}. Depending on the goal, the appropriate Markov Blanket is selected from among them.

To construct an active inference model for the interpretation of intelligent actions of living organisms or the generation of intelligent actions of artifacts, the researcher is required to design a Markov Blanket. For example, in the T-maze problem~\cite{AIF:Friston:2017}, which is typical of active inference, an active inference model is constructed from a rat, which is an agent with actions (movement in the maze). Observation variables are designed for the two modalities of the rat's exteroception (the rat's position in the maze) and interoception (attraction or aversion stimuli). In addition, hidden state variables are designed as factors that may explain the observation variables. The goal of the observation, called ``preference,'' provides a gradient of expected free energy, leading to appropriate action. Similarly, when active inference is applied to a robot, a Markov Blanket is designed starting from the robot, which is an agent with actions. Agent-centric design of Markov Blankets is useful when active inference is applied to a limited phenomenon in the world or a limited task assigned to a robot.

Agent-centric design of Markov Blanket does not address outside of the limited phenomena or tasks. Agents infer hidden states and policies defined in Markov Blanket. In turn, agents cannot infer hidden states or policies that are not defined in the Markov Blanket. Generating intelligent actions of robots and other artifacts are expected to be general-purpose, i.e., to respond to changes in the environment that are not intended by the agent. The agent's unintended changes in the environment are changes in hidden states and policies that involve the outside of the environment defined by Markov Blanket. Changes in the hidden state are caused by the presence of agents other than the target agent. For example, in the case where another robot is installed in addition to the target robot, the hidden state is changed by the position, orientation, and movement of the another robot. Changes in the policy, on the other hand, are caused by a change to a goal that the target agent cannot directly achieve. For example, in the case where the target robot performs a certain task, the policy is changed by a change to a task that cannot be performed by that robot alone. The agent does not ``intend'' these changes because they involve ``outside'' of the environment. This implies that the agent can respond by incorporating these changes ``inside'' the environment. This requires a paradigm shift in the design of the Markov Blanket from agent-centric to environment-centric.

\subsection{Environment-Centric Active Inference EC-AIF}
\begin{figure}[bt]
    \includegraphics[width=\textwidth]{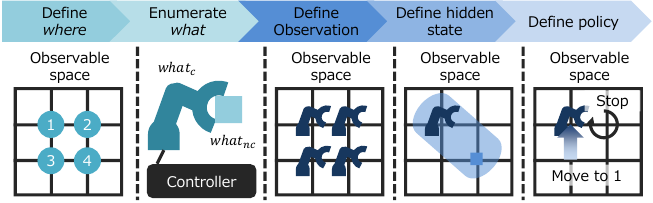}
    \caption{Decision process of observation, hidden states, and actions variables in EC-AIF.} \label{DecisionProcess}
\end{figure}

We propose an environment-centric active inference, EC-AIF, which designs Markov Blankets starting from the environment. Fig.~\ref{DecisionProcess} shows the decision process of observation variables $o$, hidden state variables $s$, and policy variables $\pi$ in EC-AIF. In EC-AIF, the $what$ and $where$ that guide modality selection in normal active inference~\cite{AIF:Parr:2022} are applied to the environment. First, $where$ is defined by considering the environment as the entire observable space. For example, the observable space is simply divided into a grid, and each grid point is defined as $where_1$, $where_2$, and so on. Then, $what$ is enumerated from the observable space, in which $where$ is independently changing. The $what$ includes $what^c$, which is connected to a controller like a robot, and $what^{nc}$, which is not connected to a controller like a ball but the $where$ changes due to robot operations. For $what^c$, those that are connected to a controller are enumerated for each controller. For $what^{nc}$, objects whose $where$ changes with robot manipulation are considered partially independent and thus included. Consequently, a ball that is glued to the robot and whose $where$ always changes with the robot is not included in $what^{nc}$. Observation variables $o$ are then defined by the possible combinations of $what$ and $where$:

\begin{equation}
    \begin{split}
        o&=\{o_{what_i} \mid i \in \mathbb{Z} _ {+}, i \leq n\} \\
        o_{what_i}&=\{where_j \mid j \in \mathbb{Z} _ {+}, j \leq m\}
        \label{eq:Observation}
    \end{split}
\end{equation}
where $n$ is the number of $what$ (including $what^c$ and $what^{nc}$) and $m$ is the number of $where$. After that, the hidden state variables $s$ are determined based on the observation variables $o$. The hidden state variables $s$ are defined by the possible combinations of observation variables $o$:

\begin{equation}
    \begin{split}
        s=\{(o_{what_{1j1}}, o_{what_{2j2}}, \ldots, o_{what_{njn}}) \mid j1, j2, \ldots, jn \in \mathbb{Z} _ {+}, j1, j2, \ldots, jn \leq m\} \label{eq:State}
    \end{split}
\end{equation}
where $n$ is the number of $what$ (including $what^c$ and $what^{nc}$) and $m$ is the number of $where$. Finally, the policy variables $\pi$ are determined based on the observation variables $o$. The policy variables $\pi$ are defined for each $what^c$ by a transition to $where$ and a stop which means not changing $where$:

\begin{equation}
    \begin{split}
        \pi=\{Move(what_i,where_j), Stop(what_i) \mid i \in \mathbb{Z} _ {+}, i \leq n, j \in \mathbb{Z} _ {+}, j \leq m\} \label{eq:Policy}
    \end{split}
\end{equation}

Algorithm~\ref{alg:FlowECAIF} shows the process flow in EC-AIF. For each $what^c$, a generative model is configured, represented by the observation $o$, hidden state $s$, and policy $\pi$ variables defined in the above decision flow. 

\begin{algorithm}[bth]
    \caption{Process flow of EC-AIF}
    \label{alg:FlowECAIF}
    \begin{algorithmic}[1]
        \FORALL{$what^c$}
            \STATE Create generative model $p(o,s,\pi)$
        \ENDFOR
        \STATE Acquire initial observation $o_0$
        \FOR{$\tau = 0$ to Timesteps $T$}
            \FORALL{$what^c$}
                \STATE Infer state $s_{\tau}$
                \STATE Infer policy $\pi_{\tau}$
                \STATE Choose action $a_{\tau}$
                \IF{$what^c$ in $a_{\tau}$}
                    \STATE Convert action $a_{\tau}$ to control value $u_{\tau}$
                    \STATE Send control value $u_{\tau}$ to controller
                    \STATE Break
                \ENDIF
            \ENDFOR
        \ENDFOR

    \end{algorithmic}
\end{algorithm}

\section{Results and Discussion}
\subsection{Experimental Setup}
\begin{figure}
    \includegraphics[width=\textwidth]{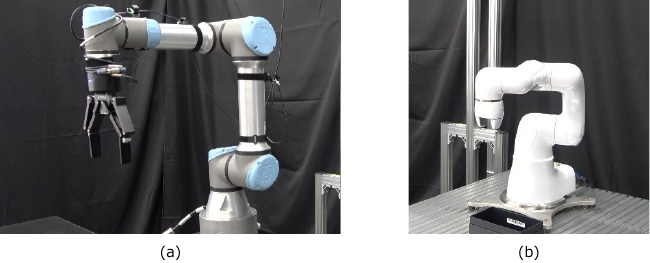}
    \caption{Robots in Experiment (a) UR5e (b) COBOTTA.} \label{Robots}
\end{figure}

The experiments were conducted with a scene of object transport by a robot. The object is mainly transferred by Universal Robots' 6-axis robot arm UR5e and Robotiq's adaptive gripper 2F-140 attached to the end of the UR5e shown in Fig.~\ref{Robots}(a). In this scene, the world is composed of the robot arm UR5e and DENSO WAVE's 6-axis robot arm COBOTTA shown in Fig.~\ref{Robots}(b), and the target object placed around UR5e. Accordingly, $what$ in this scene is as follows:

\begin{equation}
    \begin{split}
        what = \{UR5e, COBOTTA, target\ object\} \label{eq:What}
    \end{split}
\end{equation}
Furthermore, $where$ is the grid points P1 to P15, including UO, the origin position of UR5e, CO, the origin position of COBOTTA, and Int., the intermediate position between the two robots.

We evaluated the proposed method in two scenarios that capture changes in the environment in object transport:

Scenario 1: The target position for object transport changes.

Scenario 2: Another robot's orientation changes.

\begin{figure}
    \includegraphics[width=\textwidth]{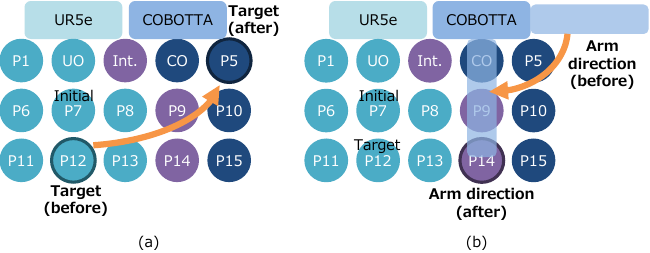}
    \caption{Scenarios of (a) the target position for object transport changes, and (b) another robot's orientation changes. Each circle represents a grid point comprising $where$. UO is the origin position of UR5e, CO is the origin position of COBOTTA, and Int. is the intermediate position between UR5e and COBOTTA. The light blue circles (P1, UO, P6, P7, P8, P11, P12, P13) are grid points in the reach range of UR5e, the blue circles (CO, P5, P10, P15) are grid points in the reach range of COBOTTA, and the purple circles (Int., P9, P14) are grid points in the reach range of both robots. In the experiment, the grid points are not evenly distributed, and the position of the points is shifted due to the constraints of the mechanism in which the robot is installed.} \label{Scenarios}
\end{figure}

The first scenario is an example of change in the environment that cannot be directly changed by the agent: the target position for object transport changes. Specifically, as shown in Fig.~\ref{Scenarios}(a), the target position is initially in front of robot arm UR5e (P12) and then changes to the side of robot arm COBOTTA (P5). Before the target position changes, the target position is within the reach of UR5e, allowing object transport by UR5e by itself. After the target position changes, the target position is outside the reach of UR5e and within the reach of COBOTTA, requiring UR5e to cooperate with COBOTTA to transport the object. The second scenario is an example of a change in the environment where there is another robot, and the target position for object transport remains the same, but the orientation of the other robot changes. Specifically, as shown in Fig.~\ref{Scenarios}(b), the target position is in front of COBOTTA (P14), and COBOTTA initially faces the other direction from the target position, and then changes to the direction of the target position. Before the orientation of COBOTTA changes, UR5e does not have contact with COBOTTA when it places an object at the target position, allowing UR5e to transport the object on its own. After the COBOTTA's orientation changes, UR5e requires cooperation with COBOTTA to transport objects because UR5e will have contact with COBOTTA when UR5e places objects at the target position. In both scenarios, the initial position of the target object was P7. The target position was given as the preference regarding the observation of the target object.

The proposed EC-AIF and the normal AIF as a benchmark are applied to both robot arms UR5e and COBOTTA. Both EC-AIF and AIF are implemented using pymdp~\cite{Heins:Pymdp:2022}, an active inference OSS. The output of the actions is passed to the robot control in the form of the target position and orientation of the robot hand. The robot control used ROS melodic, a robot OSS installed on Ubuntu 18.04.

\subsection{Change in Target Position for Object Transport}

\begin{figure}
    \includegraphics[width=\textwidth]{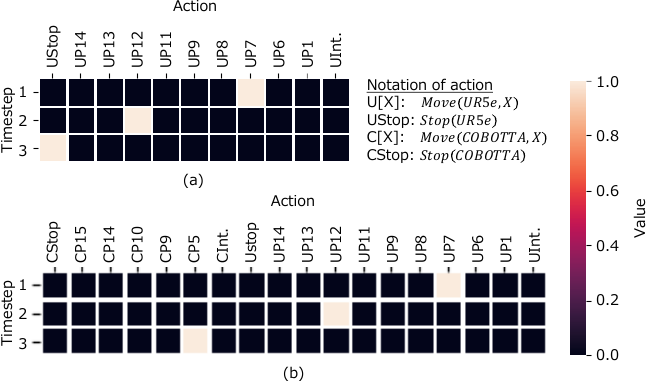}
    \caption{Transition of the selected action when the target position is within the reach of UR5e by (a) normal AIF and (b) EC-AIF.} \label{ActionWithinReach}
\end{figure}

\begin{figure}
    \includegraphics[width=\textwidth]{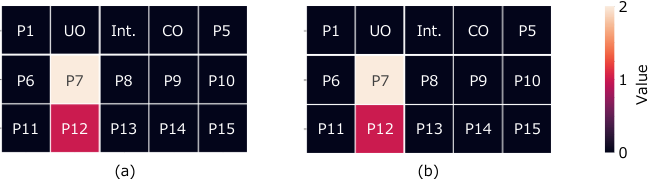}
    \caption{Total number of observations of the object when the target position is within the reach of UR5e by (a) normal AIF and (b) EC-AIF. Each cell in the heat map corresponds to $where$ of the object. The value of each cell is the total number of times the object was placed at the corresponding cell's position.} \label{ObsWithinReach}
\end{figure}

When the target position is within the reach of UR5e, the path of object transport to the target position by UR5e was obtained in both cases where normal AIF was applied and where EC-AIF was applied. Fig.~\ref{ActionWithinReach} shows the transition of the selected action when the target position is within the reach of UR5e. In both cases, in timestep 1, an action to move UR5e to P7, where the object was placed, was selected, and then in timestep 2, an action to place the object at the target position, P12 by UR5e, was selected. Fig.~\ref{ObsWithinReach} also shows the total number of observations of the object when the target position is within the reach of UR5e. In both cases, where normal AIF was applied and where EC-AIF was applied, the values of the object's starting position (P7) and P12 were relatively higher than the other positions, and a route was chosen to transport the object directly from P7 to P12.

\begin{figure}[t]
    \includegraphics[width=\textwidth]{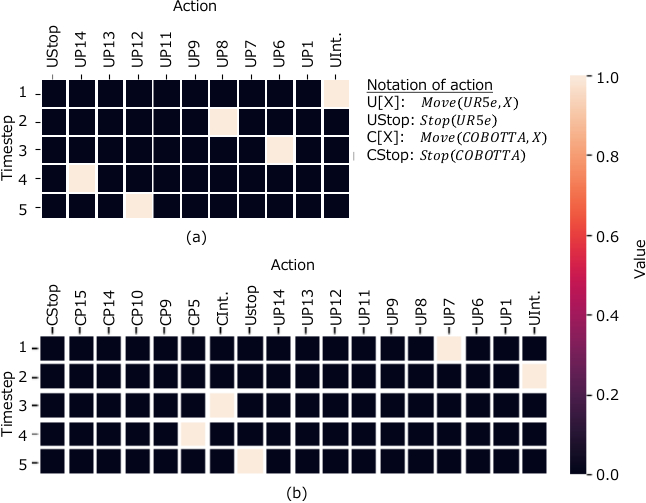}
    \caption{Transition of the selected action when the target position is outside the reach of UR5e by (a) normal AIF and (b) EC-AIF.} \label{ActionOutReach}
\end{figure}

\begin{figure}
    \includegraphics[width=\textwidth]{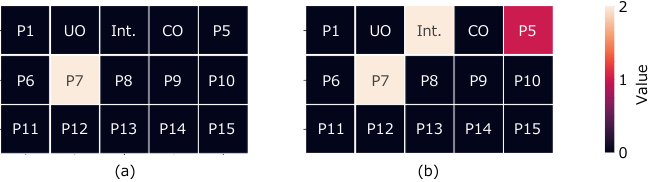}
    \caption{Total number of observations of the object when the target position is outside the reach of UR5e by (a) normal AIF and (b) EC-AIF. Each cell in the heat map corresponds to $where$ of the object. The value of each cell is the total number of times the object was placed at the corresponding cell's position.} \label{ObsOutReach}
\end{figure}

After the target position changed outside the reach of UR5e, differences were observed between cases where normal AIF was applied and cases where EC-AIF was applied. Fig.~\ref{ActionOutReach} shows the transition of the selected action when the target position is outside the reach of UR5e. In the case where normal AIF was applied, various actions of UR5e were selected at all timesteps and no consistent sequence of actions was observed. In the case where EC-AIF was applied, on the other hand, in timestep 1, the action of UR5e moving to P7, where the object was placed, was selected, followed by the action of UR5e transporting the object to the intermediate position between UR5e and COBOTTA (Int.) in timestep 2. Furthermore, in timestep 3, after the action of COBOTTA moving to the intermediate position (Int.) was selected, the action of COBOTTA placing the object to the target position, P5, was selected. Fig.~\ref{ObsOutReach} also shows the total number of observations of the object when the target position is outside the reach of UR5e. In the case where normal AIF was applied, only the value of P7, which is the initial position of the target object, is higher. In the case where EC-AIF was applied, on the other hand, the values of the object's starting position (P7), the intermediate position (Int.), and the object's target position (P5) were relatively higher than the other positions, and the shortest path was chosen for UR5e and COBOTTA to cooperatively transport from P7 to P5. Thus, the robot applying EC-AIF was able to respond to changes in target position.

The results suggest that EC-AIF can be used for environments that cannot be changed directly by the agent. In normal AIF, the environment is defined around UR5e, i.e. within the reach of UR5e. As long as the target position is within the reach of UR5e, UR5e can transport objects. In normal AIF, actions can be selected more quickly because there are fewer action variables than in EC-AIF. Once the target position is outside the reach of UR5e for some reason of the robot user, however, UR5e will not be able to determine how to transport the object. This is because the target position is now outside of the environment and no preferences regarding the object's position are defined. In EC-AIF, the environment is independent of the reach range of UR5e. Therefore, even if the target position is outside the reach range of UR5e, the actions of UR5e can be appropriately selected by only changing the preferences regarding the object's position.

The shortest path of object transport indicates that the principle of least action is implicitly included in the active inference. There are various paths other than the path selected here for object transport from the starting position P7 to the target positions P12 and P5. Nevertheless, the robot following active inference transported the object using the shortest path. The agent that follows active inference acts to minimize surprises to the environment. This means that this agent minimizes the surprise to the result of the motion of the object that follows the principle of least action. Thus, active inference naturally includes the principle of least action.

\subsection{Change in Another Robot's Orientation}

\begin{figure}
    \includegraphics[width=\textwidth]{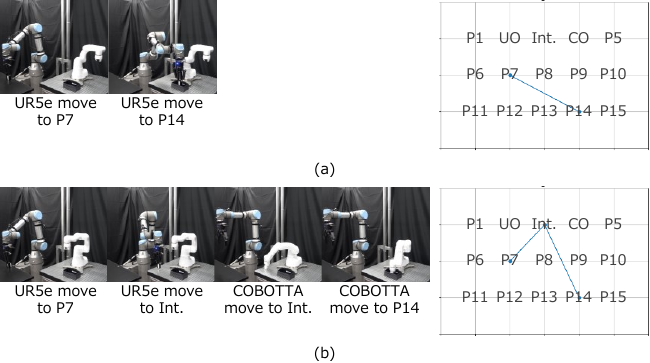}
    \caption{Motion sequences and object transport routes (a) when COBOTTA is facing in a different direction from the target position and (b) when it is facing in the direction of the target position.} \label{Trajectory}
\end{figure}

The EC-AIF allowed the object transport route to be adjusted to the direction in which COBOTTA was facing. Fig.~\ref{Trajectory} shows the motion sequences and the object transport routes when COBOTTA is facing in a different direction from the target position and when it is facing in the direction of the target position. When COBOTTA was facing the other direction from the target position, the object was directly transported from the start position to the target position. In contrast, when COBOTTA was facing the direction of the target position, the object's route changed. Specifically, the object was transported from the start position to the target position via the intermediate position (Int.). While there was no possibility of contact between UR5e and COBOTTA, the object was transported on the shortest route, and as the possibility of contact between UR5e and COBOTTA increased, the object was transported on a detour path to avoid contact.

The result suggests that the EC-AIF is capable of adapting to changes in the environment in which another robot is present. Both the starting and target positions were within the reach of UR5e. Accordingly, in a situation where there were no obstacles, including another robot, objects were transported from the starting position to the target position through the shortest path. However, when the situation changed and COBOTTA was facing the direction of the target position, COBOTTA became an obstacle in the object transport path by UR5e. As a result, even if UR5e chooses an action to transport the object to the target position, the corresponding trajectory of UR5e would not be generated. Consequently, the action via the intermediate position (Int.), which is the shortest path while avoiding the obstacle, was selected.

\section{Conclusion}
To handle unintended changes in the environment by agents, we proposed an environment-centered active inference EC-AIF that defines the Markov Blanket of active inference from the environment. In ordinary active inference, the environment is defined from the starting point of an agent that performs ``actions,'' such as a robot or a person, and the agent cannot respond to unintended changes in the environment caused by factors other than the defined environment. In the proposed EC-AIF, the environment is defined as the starting point, and there is no entity equivalent to an agent. Therefore, all states, including robots and people, are included in the inference target and unintended changes in the environment can be eliminated EC-AIF was applied to a robot arm and verified in an object transport task by a robot arm. The results showed that the robot arm successfully transported objects while responding to changes in the target position of the object and changes in the posture of other robot arms. In future work, the design of the generative model will be further refined by considering the mechanism by which each robot's preferences are transferred between robots by extending the change in the robot's trajectory due to another robot's orientation.

\bibliographystyle{splncs04}
\bibliography{reference}





\end{document}